\documentclass{article}

\PassOptionsToPackage{numbers}{natbib}
\usepackage{graphicx}
\usepackage{hyperref}
\usepackage{amsmath}
\usepackage{comment}
\usepackage{booktabs}
\usepackage{multirow}

\usepackage[preprint]{neurips_2025}

\usepackage[utf8]{inputenc} 
\usepackage[T1]{fontenc}    
\usepackage{hyperref}       
\usepackage{url}            
\usepackage{booktabs}       
\usepackage{amsfonts}       
\usepackage{nicefrac}       
\usepackage{microtype}      
\usepackage{xcolor}         

\usepackage{wrapfig}
\usepackage{booktabs}
\usepackage{siunitx}

\title{ExpertGen: Training-Free Expert Guidance \\for Controllable Text-to-Face Generation}

\author{%
  Liang Shi\\
  Northeastern University\\
  \texttt{shi.lia@northeastern.edu}
  \And
  Yun Fu\\
  Northeastern University \\
  \texttt{yunfu@ece.neu.edu}
}
\begin{document}

\maketitle

\begin{abstract}
  Recent advances in diffusion models have significantly improved text-to-face generation, but achieving fine-grained control over facial features remains a challenge. Existing methods often require training additional modules to handle specific controls such as identity, attributes, or age, making them inflexible and resource-intensive. We propose ExpertGen, a training-free framework that leverages pre-trained expert models such as face recognition, facial attribute recognition, and age estimation networks to guide generation with fine control. Our approach uses a latent consistency model to ensure realistic and in-distribution predictions at each diffusion step, enabling accurate guidance signals to effectively steer the diffusion process. We show qualitatively and quantitatively that expert models can guide the generation process with high precision, and multiple experts can collaborate to enable simultaneous control over diverse facial aspects. By allowing direct integration of off-the-shelf expert models, our method transforms any such model into a plug-and-play component for controllable face generation.
\end{abstract}

\section{Introduction}

Conditional face generation~\cite{controlnet, ip_adapter, dreamidentity, face2diffusion} is a fundamental task that aims to synthesize realistic face images based on input conditions. Conditions such as reference images, class labels, and text descriptions, enable more flexible and controllable generation for practical applications. Great progress in conditional face generation has been enabled with recent advancement of diffusion models~\cite{ddpm, ldm}. In particular, text-conditioned generation based on large foundation models~\cite{sd} now produce high-quality faces with strong alignment to the input text ~\cite{deepfakeface}. 

However, for non-text conditions, existing solutions are diverse and often rely on task-specific training. For example, ControlNet~\cite{controlnet} trains an additional input branch alongside diffusion models to enable structural information inputs like head poses and depth maps. Face personalization methods~\cite{ip_adapter, dreamidentity, face2diffusion} use facial identity as condition by training the diffusion model to accept customized ID embeddings as inputs. For each new condition to be encoded, these approaches require retraining the model with annotated data, limiting their flexibility in practice.

In this paper, we aim to eliminate the need for task-specific training for conditional text-to-face generation, and instead leverage existing face expert models in a training-free manner. 
These efforts are primarily motivated by existing research in the face analysis community, which has developed a wide range of expert models, including face recognition~\cite{arcface, adaface, insightface}, attribute classification~\cite{celeba, farl}, age estimation~\cite{mivolo}, and face segmentation~\cite{segface}. These face expert models are typically trained on large annotated datasets, and encode rich prior knowledge of facial structure and semantics. Therefore, these models are highly promising for providing guidance for face generation. 

Several existing works have explored utilizing external expert models for conditional generation. Classifier guidance~\cite{classifier_guidance} first utilizes an external classifier to steer generation in diffusion models with back-propagated gradients. More recent methods~\cite{ugd, mpgd, tfg} generalize the idea with training-free guidance, exploiting various off-the-shelf expert models to provide conditions. To compute guidance, these methods use intermediate predictions of diffusion models as inputs to the expert models. However, these predictions are inherently unrealistic and out-of-distribution for expert models trained on real images, especially in earlier denoising time steps. Complex optimization schemes were often required for compensation. 

To enable effective training-free face expert guidance, we adopt consistency models~\cite{cm} to solve above issues. Originally designed for accelerated generation, consistency models enforce a consistent mapping from noise to image in every time step. As we demonstrate, this property makes intermediate outputs significantly more realistic and in-distribution even at early stages of generation. This allows expert models to provide accurate and stable guidance throughout the sampling process. This setting contrasts from previous trained-based conditioning and naive training-free conditioning methods, and offers a reliable training-free guidance with expert supervision. 

Building on this insight, we propose ExpertGen, a training-free framework for controllable text-to-face generation that integrates any off-the-shelf face expert models into the generation process. At each time step, we feed the intermediate predictions of consistency models into selected face expert models to evaluate their misalignment with the condition, and update the predictions through back-propagation. In addition, ExpertGen introduces text-guided warmup, which complements expert guidance in earliest steps, and gradient clipping, which stabilizes guidance updates. ExpertGen demonstrates effective control over various conditions, including identity, facial attribute, age, and segmentation maps. Moreover, multiple experts can be integrated for collaborative guidance, which enables many diverse and actively studied tasks, such as face editing and age progression, to be addressed within a unified framework. By operating entirely in a training-free manner, ExpertGen turns any compatible face analysis model into a plug-and-play module for controllable face generation.

Our main contributions are as follows:

(1) We demonstrate that consistency models provide realistic intermediate predictions that are more suitable for training-free guidance with external expert models. 

(2) We propose ExpertGen, a training-free expert guidance framework for controllable text-to-face generation, enabling the use of any off-the-shelf face expert model without task-specific training.

(3) We achieve effective control across identities, attributes, ages, and segmentation maps, as well as simultaneous multi-expert guidance. We show qualitative and quantitative evidence to support the effectiveness of our guidance.  

\section{Related Work}

We briefly review relevant work in the fields of conditional text-to-face generation, training-free guidance, and consistency models. 

\subsection{Conditional Text-to-Face Generation}

In text-to-face generation, fine-grained control over facial features is often achieved by augmenting pretrained diffusion models with additional trainable modules. ControlNet~\cite{controlnet} attaches a dedicated condition encoding branch to a frozen model, enabling spatial conditioning on inputs, such as face poses and facial landmarks. IP-Adapter~\cite{ip_adapter} realizes more precise conditioning through separated cross-attention layers for image and text conditions. PreciseControl~\cite{precisecontrol} leverages features of StyleGAN to provide disentangled and continuous control. A number of methods~\cite{faceadapter,dreamidentity,face2diffusion} adopt pretrained face expert models, most commonly ID encoders, for condition encoding, which are trained to work in conjunction with the diffusion model. Most of these methods rely on training additional modules for each new conditioning, which incurs extra training cost. In contrast, we aim to implement a unified conditional face generation through training-free guidance.

\subsection{Classifier Guidance and Training-free Guidance}

Classifier guidance~\cite{classifier_guidance} is an early conditional generation method for diffusion models. It steers the generation process by using the gradient of noise-adapted classifiers, so that the generated image matches a condition defined by the classifier. More recently, training-free guidance~\cite{ugd, mpgd, freedom, tfg} attempts to circumvent the noise adaptation by using the intermediate predictions $\hat{x_0}$ as inputs of the guidance model. MPGD~\cite{mpgd} takes the gradient with respect to $\hat{x_0}$ to estimate the alignment of images and conditions, and uses an autoencoder to keep the generation within the data manifold. FreeDoM~\cite{freedom} uses a recurrence algorithm to improve image quality. UGD~\cite{ugd} further introduces multi-step optimization at each denoising step. TFG~\cite{tfg} unifies prior methods and proposes a general design space for training-free guidance. 

These advancements enable more general and flexible applications, such as low-level guidance, style-guidance, and face recognition guidance, etc. However, as intermediate predictions are still preliminary estimates of images that are far from the image distribution, these methods typically rely on complicated optimization schemes to compensate for unreliable guidance at most time steps. 

\subsection{Consistency Model}

Consistency model~\cite{cm} is a family of generative models designed to accelerate sampling by mapping noises of all levels directly to data. This nice property, referred to as self-consistency, allows consistency models to generate high-quality images in only a few steps. Further advancements of latent consistency model (LCM)~\cite{lcm} extend the idea into latent spaces and achieve state-of-the-art generation performances. LCM-LoRA~\cite{lcmlora} enables the flexibility to introduce the self-consistency property into any pre-trained diffusion models with LoRA~\cite{lora} modules. 

A particularly valuable property of consistency models is the greatly improved quality of intermediate predictions. This allows for more effective intervention of external guidance during sampling. The property has been explored for application in image personalization~\cite{lcmlookahead}, though they are primarily applied in a training-based setting. We show in this paper that with latent consistency models, a training-free guidance alone is able to facilitate high-quality conditional generation. This is especially established in the field of text-to-face generation, where a myriad of face analysis models of different functions can be collectively utilized for guidance.

\section{Methods}

We aim to achieve conditional text-to-face generation through expert guidance. We build on prior works of training-free guidance, where expert models evaluate intermediate predictions of diffusion models and back-propagate gradients to steer the generation process. 

The proposed ExpertGen is centered around providing in-distribution inputs to these expert models, enabling accurate and stable guidance without additional training. To this end, we adopt consistency models to ensure high-quality intermediate predictions throughout the diffusion trajectory. To further enhance stability, we introduce a text-guided warmup phase to avoid noisy, out-of-distribution inputs to experts, and apply gradient clipping to suppress excessively large updates during optimization.

We briefly introduce the preliminaries of latent diffusion model and consistency model in Sec.~\ref{method_preliminary}. Following that, we motivate our choice of consistency models in Sec.~\ref{method_lcm} through preliminary experiments analyzing the feature distribution of intermediate predictions. We further describe text-guided warmup in Sec.~\ref{method_warmup} and gradient clipping in Sec.~\ref{method_gradient_clipping}.

\subsection{Preliminary}
\label{method_preliminary}

The general goal of this paper is to generate realistic face images that align with a prespecified condition $c$, such as identity, age, or facial attributes. To achieve this, we build upon latent diffusion models and consistency models, which are both advanced techniques that are fundamental to high-quality image generation. 

Given a condition $c$, a time step $t$ and a noise-adding process $z_t = \sqrt{\bar{\alpha}_t} z_0 + \sqrt{1 - \bar{\alpha}_t} \epsilon$, a diffusion model $\epsilon_\theta$ is trained to denoise by predicting the added Gaussian noise from noisy representation, minimizing the objective:
\begin{equation}
\mathbb{E}_{t, z_t, \epsilon} \|\epsilon - \epsilon_\theta(z_t, t, c)\|^2.
\end{equation}
This results in a model that is capable of generating high quality images by iteratively denoising from a Gaussian noise to a clean image representation. In each denoising step, we obtain an estimated prediction with
\begin{equation}
\label{z0}
f_\theta(z_t, t, c) = \tilde{z}_{0|t} =  \frac{1}{\sqrt{\bar{\alpha}_t}} \left(z_t - \sqrt{1 - \bar{\alpha}_t} \cdot \epsilon_\theta(z_t, t, c)\right).
\end{equation}
A deterministic DDIM~\cite{ddim} sampling step directly utilizes this estimation to compute the latent at a future denoising step $T=s$ to continue the denoising process with $z_{s} = \sqrt{\bar{\alpha}_s} \tilde{z}_{0|t} + \sqrt{1 - \bar{\alpha}_s} \cdot \epsilon.$
For latent diffusion models~\cite{ldm}, a decoder maps the latent back to the pixel space, producing either the final generated image or an intermediate result through $\hat{x}_{0|t} = D(\hat{z}_{0|t})$. Throughout this paper, we refer to this $\hat{x}_{0|t}$ as the intermediate predictions to be fed to the expert models. 

Since the introduction of diffusion models, various accelerated variants have been proposed. Among them, consistency models~\cite{cm, lcm} learn to produce consistent denoised predictions at varying time steps, and are capable of generating high-quality predictions in a few steps. Consistency models are typically distilled from pre-trained diffusion models with the objective of
\begin{equation}
\mathbb{E}_{t, z_t, \epsilon} \|f_\theta(z_t, t, c) - f_{\theta^-}(z_{t+k}, t+k, c) \|^2,
\end{equation}
where $f_{\theta^-}$ is the distillation target and $k$ determines a random offset to another time step. As we demonstrate, the self-consistency property of consistency models not only accelerates sampling, but also produces cleaner intermediate predictions that can be leveraged for improving expert guidance. 

\subsection{Expert Guidance with Consistency Models}
\label{method_lcm}

In order to introduce training-free guidance to denoising process, existing methods~\cite{ugd, tfg, freedom} exploit the intermediate prediction $x_{0|t}$, as it serves as an estimate of the clean image at each time step. An expert-driven loss is evaluated based on this prediction, and its gradient is back-propagated to steer the generation process. The denoising objective is then modified as follows:
\begin{equation}
\label{update_denoising}
 \bar{\epsilon_\theta}(z_t, t) = \epsilon_\theta(z_t, t) - w\sqrt{1 - \bar{\alpha}_t}\nabla_{z_t} L_{exp}(\hat{x}_{0|t}^{LDM}),
\end{equation}
where $L_{exp}$ defines a loss specified by an external expert model, and $w\sqrt{1 - \bar{\alpha}_t}$ serves as a weighting parameter to balance the denoising objective and external objectives. 

To better understand the suitability of intermediate predictions for expert guidance, we conduct a proof-of-concept experiment centered on face images. We motivate the idea of introducing consistency models into the framework by comparing the face generation dynamics of latent diffusion models (LDMs) and latent consistency models (LCMs).

\begin{figure}[t]
    \centering
    \includegraphics[width=1\textwidth]{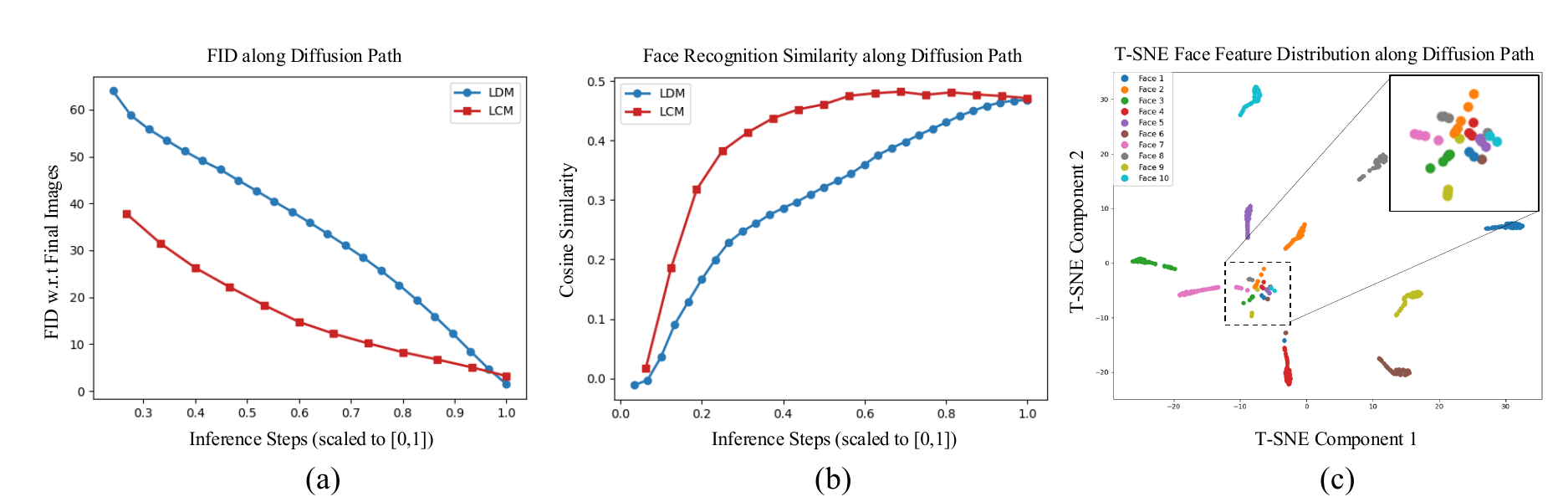}
    \caption{\textbf{Evaluating the quality of intermediate predictions. }(a) FID of face images at different DDIM steps (earliest steps omitted due to large values). While both models improve, LCM achieves higher image quality much faster. (b) Face recognition features of LCM’s intermediate predictions converge within early steps. (c) Low-quality early-step predictions lack distinctive facial features, and therefore form a multi-color cluster in feature space, resulting in ambiguous gradient guidance. }
    \label{fig:motif}
    \vspace{-1em}
\end{figure}

We generate 100 face images respectively with LCM and LDM, and collect the intermediate predictions along the diffusion path. In terms of image quality measured by FID in Fig.~\ref{fig:motif}(a), LCMs exhibit a much more progressive improvement: by only 30\% of the total timesteps, LCMs already produce images comparable in quality to those generated by LDMs at 70\% of the timesteps. Similarly, in Fig.~\ref{fig:motif}(b), we observe that LCMs recover fine-grained facial details much earlier in the generation process. This is reflected in the rapid increase in cosine similarity between the face embeddings of intermediate predictions and a referenced ground truth image.

We further study how face analysis models interpret early denoising outputs. We collect every intermediate prediction for ten identities along a 30-step DDIM trajectory from LDMs, embed them with a face representation model~\cite{farl}, and project the features with t-SNE~\cite{tsne}, presented in Fig.~\ref{fig:motif}(c). While the model is robust enough to cluster most features by identity, note how a multi-color cluster exists in the center, indicating that the representation model assigns many early-step images to ambiguous regions of the feature space. In such cases, expert models cannot extract reliable information, which leads to noisy or even misleading expert guidance.

These results are consistent with our intuition. While LDM-generated faces induce unreliable guidance, LCM-generated faces become clear and detailed at much earlier stages, which provides reliable inputs for expert models and enables accurate predictions and guidance in the generation process. Therefore, ExpertGen preserves the existing training-free guidance formulation in Eq.~\ref{update_denoising} but integrates LCMs to produce high-quality intermediate predictions in the denoising process, enhancing the effectiveness of expert guidance.

\subsection{Text-guided Warmup}
\label{method_warmup}

Replacing LDMs with LCMs significantly improves the overall quality of predicted faces and shrinks the size of the ambiguous image cluster in Fig.~\ref{fig:motif}(c). However, images from the earliest few steps of denoising inevitably remain blurry and incomprehensible to both human eyes and external expert models. We aim to entirely avoid feeding these out-of-distribution images into expert models by introducing a simple warmup phase to the generation process. For the early denoising steps of $T>T_{thre}$, we disable the expert guidance by setting the weight to zero, and replace it by text conditions that explicitly describes the target attributes:
\begin{quote}
\texttt{A realistic photo of a {CONDITION} face looking at camera. }
\end{quote}
For attributes describable in text (e.g., age, facial attributes), we directly insert the desired condition into the prompt. For tasks like identity and segmentation map guidance, we enrich the prompt with descriptions of facial attributes extracted from the reference image. 

Although these text prompts offer complementary information, we empirically show that text alone fails to provide sufficient guidance for condition-aligned generation, especially for smaller models. However, this design allows the text to guide the model when expert feedback is unreliable, with expert models taking over as the predictions become more detailed and in-distribution. We demonstrate that ExpertGen built on this text-guided warmup phase significantly outperforms generation with the above enriched text conditions alone.

\subsection{Gradient Clipping}
\label{method_gradient_clipping}

We observe significant fluctuations in expert model predictions along the diffusion trajectory. These fluctuations often result in excessively large gradient values, which frequently cause generation failures, typically in the forms of low-quality images. To mitigate these negative effects, we perform an element-wise clipping on the calculated gradient to stabilize the expert gradient: 
\begin{equation}
\label{final_eq}
 \bar{\epsilon_\theta}(z_t, t) = \epsilon_\theta(z_t, t) - w\sqrt{1 - \bar{\alpha}_t}\cdot \mathrm{clip}(\nabla_{z_t} L_{exp}(\hat{x}_{0|t}^{LCM}), -\tau, \tau),
\end{equation}
Empirical results indicate that this operation preserves guidance effectiveness while significantly improving image quality. As a result, it remains a crucial component of our design.

\section{Experiments}

We evaluate ExpertGen across diverse conditional face generation tasks. Implementation details, including expert model choices and hyper-parameters, are in Sec.~\ref{exp:details}. Single-expert results are shown in Sec.~\ref{exp:single}, multi-expert guidance in Sec.~\ref{exp:multi}. Ablations of our design choices and evaluations  are presented in Sec.~\ref{exp:ablation}.

\begin{table}[h]
\centering
\caption{Summary of tasks, guidance loss functions, evaluation metrics, and penalties for failed face detections. $\hat{p}$ and $\hat{p}_i$ denote predicted probabilities, $\hat{y}$ and $\hat{y}_i$ denote predicted labels, while $y$, $y_i$, $a_{\text{gt}}$, and $\mathbf{e}_{\text{gt}}$ refer to conditioning targets.}
\begin{tabular}{lccc}

\toprule
\textbf{Task} & \textbf{Loss Type} & \textbf{Guidance Loss $L_{exp}$} & \textbf{Evaluation Metric} \\
\midrule
Identity & Embedding Similarity & $1 - \cos(\hat{\mathbf{e}}, \mathbf{e}_{\text{gt}})$ & $\cos(\hat{\mathbf{e}}, \mathbf{e}_{\text{gt}})$ \\
Attribute & Classification & $- y \log \hat{p}$ & $\mathbb{I}(\hat{y} = y)$ \\
Age & Regression & $|\hat{a} - a_{\text{gt}}|$ & $|\hat{a} - a_{\text{gt}}|$ \\
SegMap & Dense Prediction & $- \sum_{i} y_i \log \hat{p}_i$ & $\frac{1}{P} \sum_{i=1}^{P} \mathbb{I}(\hat{y}_i = y_i)$  \\
\bottomrule\\
\end{tabular}
\vspace{-2em}

\label{tab:guidance_tasks}
\end{table}

\subsection{Implementation Details. }
\label{exp:details}

We primarily experiment with Stable Diffusion v1.5 (SD-v1.5)~\cite{sd} for consistency with prior works. Additionally, we provide results on SDXL~\cite{sdxl} to demonstrate the generalizability to stronger generative models. These models are equipped with LCM-LoRA adapters to enable consistency-based sampling, following~\cite{lcmlora}. By default, we introduce expert guidance at time steps $T_{thre}=800$ after the warmup phase. Guidance scales are set at $w=200$, and gradient clipping thresholds are set at $\tau=5e-4$. Images are generated with 16 denoising steps by default, though we show that some guidance tasks can be solved in 8 steps. All guidance experiments can be performed on a single NVIDIA A6000 GPU. 

Four types of conditions are considered: identity, facial attribute, age, and segmentation map. These tasks correspond to different types of losses, including embedding similarity, classification, regression, and dense prediction. The specific task types, guidance loss functions, and evaluation metrics are summarized in Tab.~\ref{tab:guidance_tasks}. We choose Arcface~\cite{arcface} for ID guidance, FaRL~\cite{farl} for facial attribute guidance, MiVOLO~\cite{mivolo} for age guidance, and Segface~\cite{segface} for segmentation map guidance. Guidance targets for ID and segmentation are selected from VGGFace2~\cite{vggface2}.   Attribute descriptions, when involved, are obtained with FaRL~\cite{farl}. While we strive to select state-of-the-art expert models of each face analysis task, these experts can be replaced with any off-the-shelf alternatives.

We use task-specific expert models to evaluate the alignment between the target and the generated image. For generated images where no face is detected, we assign a penalty error to reflect failure in conditioning.  Additionally, we evaluate the visual quality of generated images with FaceScore~\cite{facescore}, which is designed for evaluating image aesthetics and human preferences specifically in the context of text-to-face generation. 

\begin{table}[h]
\centering
\caption{Comparison of training-free guidance methods on conditions of identity, facial attribute, age, and segmentation maps. We compare with generation with no guidance, text-only guidance, LDM guidance. Specific metrics are detailed in Tab.~\ref{tab:guidance_tasks}.}
\begin{tabular}{lcccc}

\toprule
 & {Identity ($\uparrow$)} & {Attribute ($\uparrow$)} & {Age ($\downarrow$)} & {SegMap ($\uparrow$)} \\
\midrule
No guidance & 0.000 & 0.230 & 16.211 & 0.618 \\
Text guidance & 0.002 & 0.490 & 4.930 & 0.695 \\
LDM Guidance (UGD~\cite{ugd}) & 0.181 & 0.625 & 6.144 & 0.461 \\
\textbf{ExpertGen (Ours)} & \textbf{0.594} & \textbf{0.808} & \textbf{1.830} & \textbf{0.768} \\
\bottomrule\\
\end{tabular}
\vspace{-2em}
\label{tab:guidance_results}
\end{table}

\subsection{Single Expert Guidance}
\label{exp:single}

Table~\ref{tab:guidance_results} presents quantitative results of single-expert guidance, compared against random generation, text-conditioned generation, and LDM-based training-free guidance. We implement this guidance with UGD~\cite{ugd}, as it offers a solution in the context of latent diffusion models, whereas other existing work focus on domain-specific models. We show that adding the condition-specifying text guidance and existing LDM guidance improve the result in most cases, though our ExpertGen achieves the best performance across all tasks. In the following, we provide detailed analysis and qualitative results for each task separately.

\begin{figure}[h]
    \centering
    \includegraphics[width=1\textwidth]{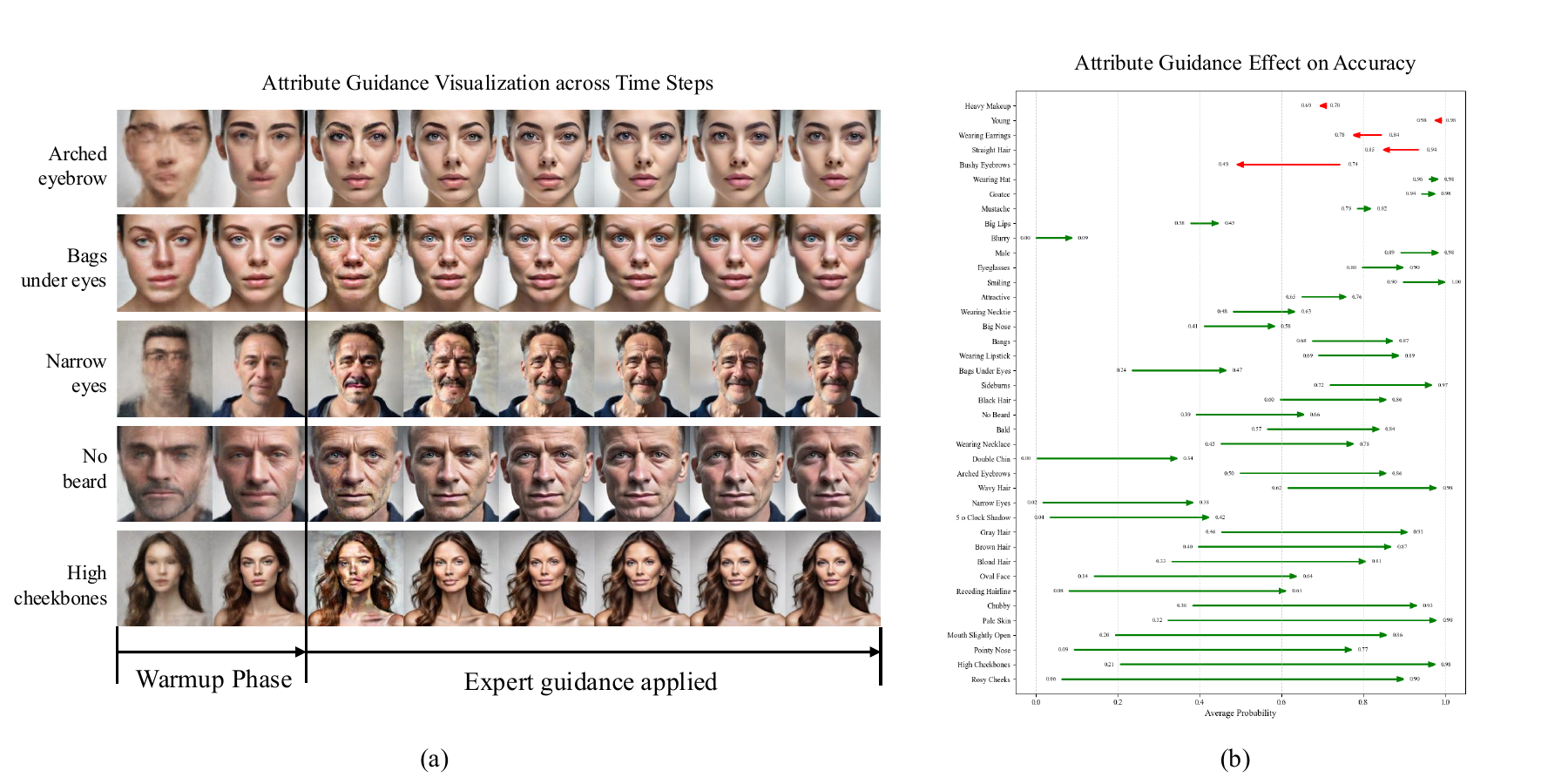}
    \vspace{-2em}
    \caption{\textbf{Qualitative and quantitative results of facial attribute guidance.} (a) We select five attributes that are challenging to generate using text conditions alone, and demonstrate how ExpertGen effectively morphs the image across eight DDIM time steps to generate correct attributes (zoom in for details). (b) Average probability of successful generation before and after ExpertGen across all 40 Celeb-A~\cite{celeba} facial attributes. Most attributes gain substantial improvements with ExpertGen. }
    \label{fig:attr}
    \vspace{-2em}
\end{figure}

\textbf{Attribute Guidance.} We select random attributes from the CelebA~\cite{celeba} dataset as condition and measure the accuracy of generated images. With ExpertGen, the generated images align with required facial attribute conditions with an accuracy of 80.8\%, surpassing pure text and LDM guidance. 

In Fig.~\ref{fig:attr}(a), we showcase five attributes that are challenging for models to generate with text conditions alone. We present the intermediate predictions of LCMs along the diffusion path, where models initialize the generation with text guidance and later involve expert guidance. It is evident that after the warmup phase, the LCM produces more in-distribution samples, enabling expert models to provide meaningful gradients. Although the generated images may remain incorrect at the end of warmup, we observe that the relevant facial regions begin to morph as expert guidance is introduced. These noisy artifacts are gradually refined, with the correct attributes emerging by the end of the denoising process. For instance, the “No beard” text condition initially fails by producing a beard (a common limitation of generative models), but ExpertGen successfully removes it in subsequent steps.

We provide the average probability of correct attribute generation with and without ExpertGen in Fig.~\ref{fig:attr}(b). While some attributes, such as "male", are generated reliably using text conditions alone, the majority of the 40 attributes are not, with about half exhibiting average probabilities below 0.5. Our method substantially improves their success rates across most attributes, notably with “rosy cheeks” improving from 0.06 to 0.90. In the Appendix, we show a similar figure for the attribute guidance effect on SDXL, where the generation of most attributes are improved to near-perfect levels.

\textbf{Age Guidance.} We show that the age condition can be modeled reasonably well with text guidance alone, producing an average age error of 5 years. LDM methods introduce excessive noise to the image, causing lower image quality and higher errors. Our method further refines the conditioning, reducing the error to 1.83 years. We note that this is even below the average error of state-of-the-art age estimator on real images, which limits the reliability of absolute errors. Still, the sharp reduction promises the potential of ExpertGen to align with future improved versions of age estimator.

\begin{figure}[h]
    \centering
    \includegraphics[width=1\textwidth]{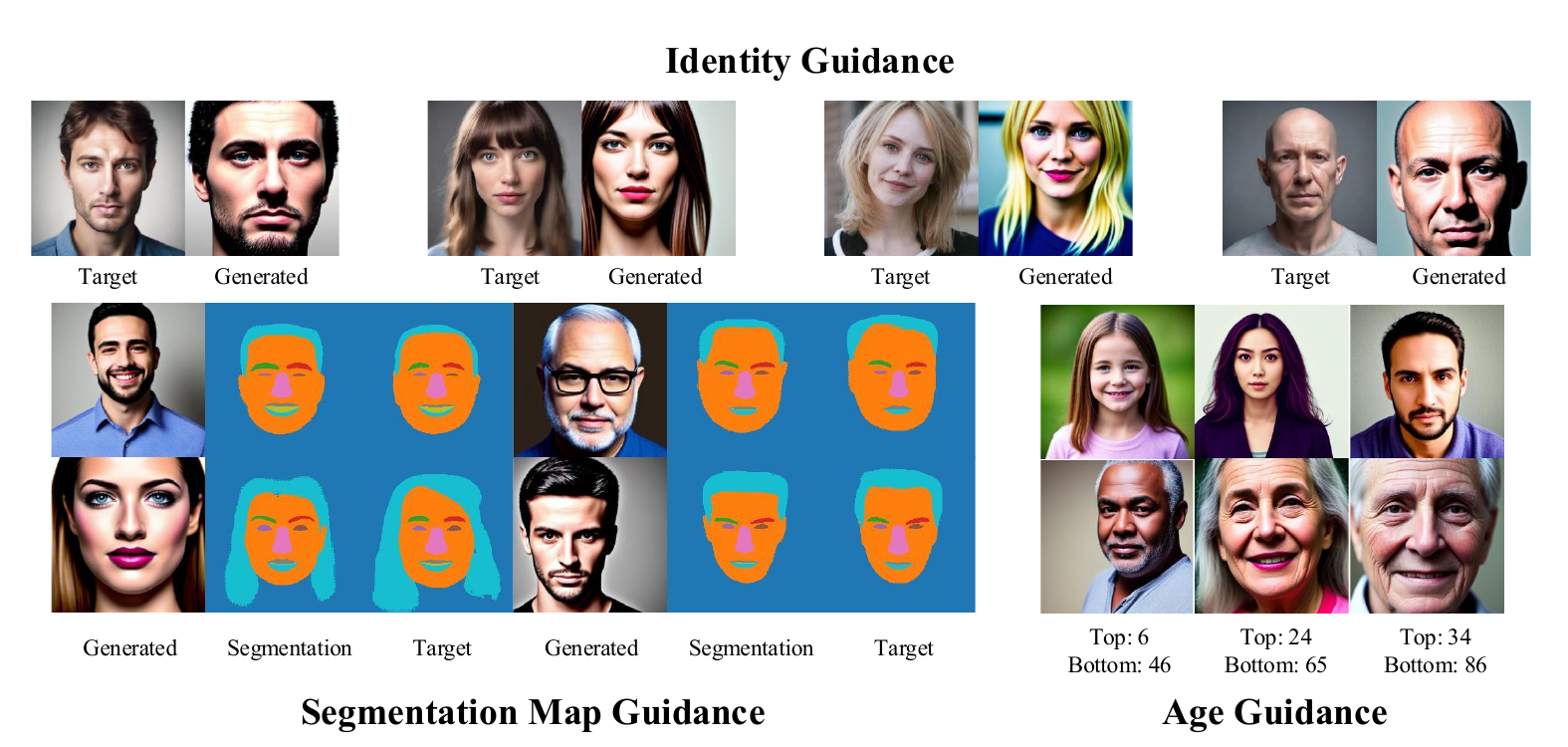}
    \vspace{-2em}
    \caption{Visualizations of ID guidance, segmentation map guidance, and age guidance. Conditions of each task is provided next to the generations: a reference image for ID guidance, a target segmentation map for segmentation guidance, and an age label for age guidance. }
    \label{fig:three-guidance}
\end{figure}

\begin{wraptable}{r}{0.5\linewidth}
\centering
\small
\renewcommand{\arraystretch}{0.9}
\caption{Comparison of ID Guidance with ID-preserved generation methods. We report results under comparable experimental settings.}
\begin{tabular}{l c}
\toprule
Method & ID Similarity \\
\midrule
DreamIdentity & 0.467 \\
Face2Diffusion & 0.363 \\
RectifiedID & 0.486 \\
\textbf{ExpertGen (Ours)} & \textbf{0.516} \\
\bottomrule
\end{tabular}
\vspace{-1em}

\label{tab:id_similarity}
\end{wraptable}

\textbf{ID Guidance.} The objective of ID guidance is to generate a face image aligned with the identity of a reference image, in line with existing ID-preserved generation methods~\cite{dreamidentity, face2diffusion, rectifid}. As a training-free approach, ExpertGen achieves comparable or better face similarity to these methods, highlighting its versatility in performing diverse conditional generation tasks.

\textbf{Segmentation Map Guidance.} We obtain segmentation maps from a reference image using face parsing tools~\cite{segface} and guide the model to generate faces with matching segmentation maps. As faces are aligned for parsing, all groups achieve reasonably well pixel accuracy. Nevertheless, our method improves the accuracy from 69.5\% to 76.8\%. This extends the applicability of our approach to spatially conditioned generation tasks, which is a strength of methods like ControlNet~\cite{controlnet}. We provide visualizations of age guidance, ID guidance and segmentation map guidance in Fig.~\ref{fig:three-guidance}. 

\begin{figure}[h]
\vspace{-0.5em}
    \centering
    \includegraphics[width=1\textwidth]{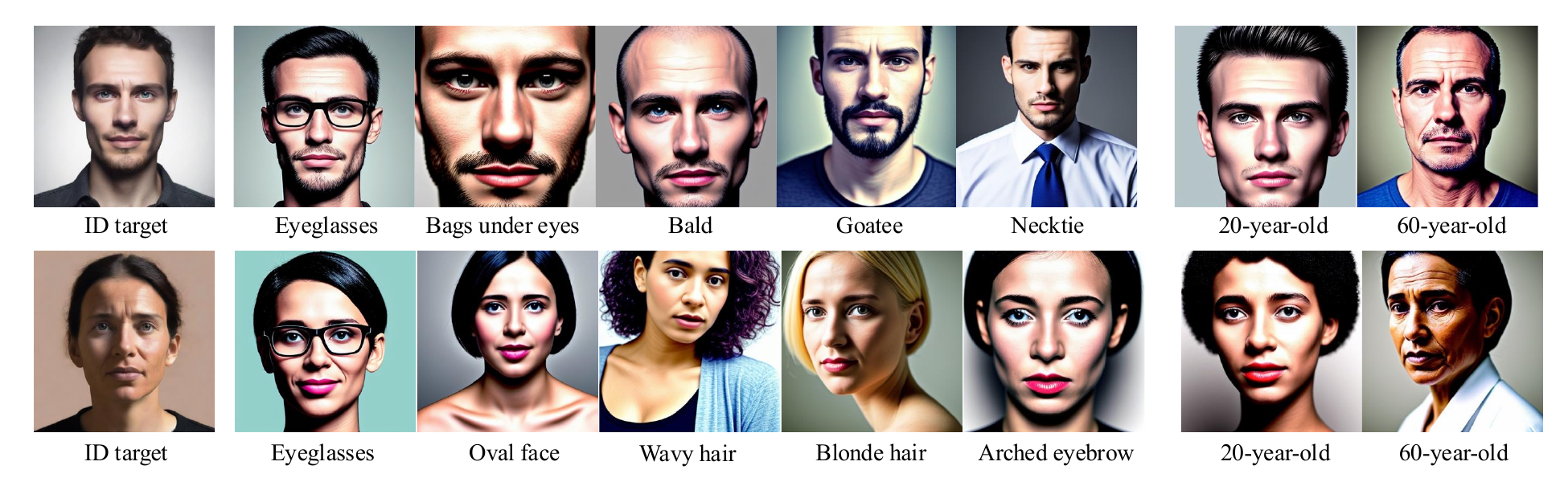}
    \caption{Visualizations of multi-expert guidance. Given an ID target on the left, we simultaneously apply ID and attribute or age guidance to generate images of the same person under new conditions. }
    \label{fig:multi}
    \vspace{-1em}
\end{figure}

\subsection{Multi-Expert Guidance}
\label{exp:multi}

We demonstrate the versatility of ExpertGen by incorporating guidance from multiple experts simultaneously, such as identity and age, or identity and attributes. To achieve this, we simply let the intermediate predictions enter multiple expert networks, and add up the back-propagated gradients. In Fig.~\ref{fig:multi}, we show that both the ID target and specific attributes or age labels are correctly followed in the generated images. Similar objectives have been explored in identity-preserving generation~\cite{dreamidentity, face2diffusion, ip_adapter} and other face editing research, but our method achieves them without any task-specific training. 

\subsection{Ablation Study}
\label{exp:ablation}
\vspace{-0.5em}

We conduct ablation studies for crucial components in our research, especially the contribution of LCM guidance, text-guided warmup phase, and gradient clipping. We also discuss hyper-parameter choices of clipping threshold $\epsilon$, guidance weight $w$, and warmup length ($T_{thre}$), with details presented in the Appendix. Finally, we demonstrate an invariance of our results to different evaluation models.

\begin{table}[t]
\centering
\scriptsize
\renewcommand{\arraystretch}{1.05}

\vspace{-2em}

\begin{minipage}{0.60\linewidth}
\centering
\caption{Ablation study on the contributions of text-guided warmup (Text), LCM guidance (LCM), and gradient clipping (Clip). We report accuracy of generated attributes for guidance effective, and FaceScore for quality.}
\begin{tabular}{ccc|cc|cc}
\toprule
\textbf{Text} & \textbf{LCM} & \textbf{Clip} & \multicolumn{2}{c|}{\textbf{SD-v1.5}} & \multicolumn{2}{c}{\textbf{SDXL}} \\
 & & & Accuracy & Quality & Accuracy & Quality \\
\midrule
\checkmark & & & 0.483 & 4.342 & \underline{0.720} & \textbf{4.994} \\
 & \checkmark & & 0.573 & \textbf{4.554} & 0.540 & 4.178 \\
\checkmark & \checkmark & & \underline{0.785} & 3.711 & 0.703 & 3.997 \\
\checkmark & \checkmark & \checkmark & \textbf{0.808} & \underline{4.387} & \textbf{0.853} & \underline{4.983} \\
\bottomrule
\end{tabular}
\label{tab:ablation_main}
\end{minipage}
\hfill
\begin{minipage}{0.38\linewidth}
\centering


\begin{minipage}{\linewidth}
\centering

\caption{Effect of guidance weight $w$}
\begin{tabular}{ccc}

\toprule
\textbf{w} & Accuracy & Quality \\
\midrule
100 & 0.840 & \textbf{4.967} \\
200 & \textbf{0.841} & 4.812 \\
300 & 0.812 & 4.578 \\
\bottomrule
\end{tabular}
\label{tab:ablation_weight}

\end{minipage}

\vspace{2mm}


\begin{minipage}{\linewidth}
\centering
\caption{Effect of clipping threshold $\tau$}
\begin{tabular}{ccc}
\toprule
\textbf{$\tau$} & Accuracy & Quality \\
\midrule
0.002  & 0.844 & 4.688 \\
0.001  & \textbf{0.856} & 4.890 \\
0.0005 & 0.853 & \textbf{4.983} \\
\bottomrule
\end{tabular}
\label{tab:ablation_clip}
\end{minipage}
\end{minipage}

\vspace{-2em}
\end{table}

\subsubsection{Basic components and hyper-parameters}

We report the performance of our method on both SD-v1.5 and SDXL with core components ablated in Tab.~\ref{tab:ablation_main}. We focus on the attribute guidance task, and report attribute accuracy along with image quality, measured by FaceScore~\cite{facescore}. 

Our text-guided warmup phase inserts a target-specifying prompt into the model, allowing both variants to achieve decent accuracy. Nevertheless, both models retain room for improvement, especially on challenging attributes as detailed above.

Expert models that receive in-distribution inputs produced by LCMs improve accuracy for SD-v1.5. However, larger guidance weight does not necessarily improve success rates, and often degrades image quality (Tab.~\ref{tab:ablation_weight}). We also note that the delayed LCM guidance without text warmup (second row in Tab.~\ref{tab:ablation_main}) does not yield promising results and even degrades accuracy on SDXL. This is likely because the unconditional generation before expert guidance is highly misaligned with the target condition, leading to large and unstable gradients. This highlights the importance of first approximating the condition via text-guided warmup in enabling successful guidance.

Finally, gradient clipping plays a major role in preserving image quality, while also improving accuracy. As shown in Tab.~\ref{tab:ablation_clip}, aggressive clipping maintains accuracy while substantially improving image quality. All components jointly contribute to improving attribute generation accuracy to 80.8\% and 85.3\% on SD-v1.5 and SDXL respectively, substantially outperforming the baseline.

\subsubsection{Invariance to evaluation models}

A common concern for training-free guidance is its potential adversarial effect~\cite{robust_classifier_guidance}, which can cause the guiding expert model to favor the generated result. We show that different expert models respond positively to our generated images, indicating that our method does not induce significant adversarial behavior. Detailed experimental evidence is provided in the Appendix.

\section{Limitations} 
\vspace{-0.5em}
While our method offers a general, training-free solution for text-to-face generation, its performance depends on the quality of the expert models used. Particularly, two practical challenges remain:

\textbf{Spurious correlations. } The expert models potentially take advantage of spurious correlations in achieving superior discriminative abilities. For instance, guiding the attribute “blonde hair” often biases the generation toward female faces. Incorporating appropriate text conditions and multi-objective guidance mitigate such unintended effects.

\textbf{Limitation of experts. } The framework inherits the limitations of the expert models it relies on. As noted earlier, one example is the less accurate current age models, which sets a lower bound on the achievable accuracy. Nevertheless, the plug-and-play property of ExpertGen allows future expert models to be easily integrated, improving generation quality as they advance.
\vspace{-0.5em}

\section{Conclusion}
\vspace{-0.5em}

Modern text-to-image models rely heavily on large-scale text-image datasets for supervision. However, they struggle with conditions that are inherently difficult to describe with text, such as facial identities, or conditions that suffer from limited paired data, such as fine-grained facial attributes. In contrast, knowledge of these complex visual concepts is embedded in domain-specific expert models, which are trained with diverse objectives and datasets to achieve precise discrimination. 

In light of this, we present ExpertGen, a general framework that augments text-to-face generation with guidance driven by these experts, which addresses the limitations of general-purpose models in a training-free manner. This has been achieved by combining consistency models and carefully designed complementary modules to provide detailed, in-distribution predictions to expert models. We hope this work encourages further exploration of cost-efficient methods for incorporating non-textual, domain-specific knowledge for conditional face and general image generation.

\bibliographystyle{plainnat}
\setlength{\bibsep}{4pt plus 0.3ex}
\bibliography{neurips_2025}

\newpage
\appendix

\section{Additional Results}

\subsection{Additional ablation study}

We present additional results on hyper-parameter settings, including warmup threshold $T_{thre}$, clipping threshold $\tau$, and guidance weight $w$. All results are obtained under the setting of facial attribute guidance for SDXL, and images are generated with 8 steps. 

\textbf{Length of text-guided warmup.} Text-guided warmup is a crucial component in our design. We delay the introduction of expert guidance and denoise with enriched text prompts, which ensures that the experts accept high-quality images produced in later steps. We study the optimal time step to begin expert guidance, striking a balance between high-quality expert inputs and guidance effectiveness. 

We adopt an 8-step DDIM schedule and vary the introduction of expert guidance from the first to the sixth step. These correspond to $T_{thre}$ values decreasing from 1000 to 500 with a gap of 100, and step 3 corresponds to our default setting of $T_{thre}=800$. We compare these settings to text-only guidance, which provides the highest image quality but serves as a baseline in attribute accuracy. 

Our results in Tab.~\ref{supp-tab-warmup} show that introducing expert guidance at the earliest steps degrades image quality, a result of inaccurate gradient signal from out-of-distribution expert inputs. The quality improves and returns to baseline when the guidance begins at the third step or later. However, delaying guidance further reduces attribute accuracy, as the scaled gradients become too weak to effectively steer the generation. These results reinforce the necessity of designing different phases combining text guidance and expert guidance, and underscore the importance of timely expert guidance introduction.

\begin{table}[h]
\centering
\caption{Effect of varying the text-guided warmup length, measured by the DDIM step at which expert guidance begins, under an 8-step DDIM schedule. \textbf{Bold} text corresponds to the optimal step.}
\begin{tabular}{lccccccc}
\toprule
 Steps  & 1 & 2 & 3 & 4 & 5 & 6 & Text-only\\
\midrule
Accuracy  & 0.790 & 0.831 & \textbf{0.853} & 0.828 & 0.814 & 0.753 & 0.720\\
Quality   & 4.396 & 4.847 & \textbf{4.983} & 5.022 & 5.084 & 5.066 & 4.994\\
\bottomrule
\end{tabular}
\label{supp-tab-warmup}
\end{table}

\textbf{Sweeping gradient clipping thresholds and guidance weights.} The distribution of our final guidance signal $w\sqrt{1 - \bar{\alpha}_t}\cdot \mathrm{clip}(\nabla_{z_t} L_{exp}(\hat{x}_0), -\tau, \tau)$, as shown in Eq. \ref{final_eq}, is jointly determined by two hyper-parameters. The clipping threshold $\tau$ directly bounds the range of gradient values, while $w$ scales their overall magnitude. Together, they control both the amplitude and sparsity of the expert guidance. We conduct a hyperparameter sweep presented in Tab.~\ref{tab:sweep} to identify the optimal pair and examine their interaction with generation performance.

\begin{table}[h]
\centering
\caption{Effect of clipping threshold $\tau$ and guidance weight $w$ on accuracy and quality. \textbf{Bold} text corresponds to the optimal pair we select.}
\label{tab:sweep}
\begin{tabular}{ccccccc}
\toprule
\multirow{2}{*}{Clipping threshold $\tau$} & \multicolumn{2}{c}{$w=100$} & \multicolumn{2}{c}{$w=200$} & \multicolumn{2}{c}{$w=300$} \\
\cmidrule(r){2-3} \cmidrule(r){4-5} \cmidrule(r){6-7}
& Accuracy & Quality & Accuracy & Quality & Accuracy & Quality \\
\midrule
0.0005 & 0.837 & 4.991 & \textbf{0.853} & \textbf{4.983} & 0.846 & 4.931 \\
0.001  & 0.834 & 4.966 & 0.856 & 4.890 & 0.824 & 4.788 \\
0.002  & 0.840 & 4.967 & 0.844 & 4.688 & 0.809 & 4.524 \\
\bottomrule
\end{tabular}

\end{table}

Our general finding is that a constrained guidance signal provides the best trade-off between accuracy and image quality. While increasing guidance weight $w$ moderately improves accuracy across different threshold $\tau$, a weight of 300 reduces both accuracy and quality. This suggests that larger gradient values harm the balance between the denoising objective and the guidance, which disrupts the denoising trajectory. Similarly, at each fixed $w$, increasing the threshold $\tau$ consistently degrades image quality, and even harms accuracy at $w=300$. As $\tau$ increases, the effect of gradient clipping diminishes, allowing unstable gradients to dominate, which leads to noticeable quality degradation. In contrast, a strong clipping of $\tau$ = 0.0005 preserves image quality without hurting accuracy, indicating that expert gradients remain informative even under tight constraints.

These results suggest that moderate guidance weights and tight clipping produces the best result for ExpertGen. Therefore, we select $w=200$ and $\tau=0.0005$ for our default setting. Slight tuning of these parameters may benefit individual tasks, though the trends observed in this table are broadly instructional and apply consistently across tasks.

\subsection{Varying evaluation networks}
Since the guidance network is frozen and not jointly trained with the generative model, the generator may over-optimize toward the guidance objective. This can lead to adversarial samples, which satisfy the target constraint from the perspective of the guidance model, but appear misaligned with conditions to human viewers. 

\begin{table}[h]
\centering
\caption{Identity similarity scores under different face recognition methods.}
\begin{tabular}{lccc}
\toprule
Method & ArcFace & Buffalo-l & Antelope-v2 \\
\midrule
Text Guidance & 0.020 & 0.010 & 0.015 \\
UGD & 0.181 & 0.171 & 0.193 \\
ExpertGen & 0.594 & 0.473 & 0.516 \\
\bottomrule
\end{tabular}
\label{tab:id_guidance}
\end{table}

We examine whether changing evaluation networks results in significant fluctuations. In Tab.~\ref{tab:id_guidance}, we demonstrate that this does not pose an issue in our framework. In the ID guidance setting, we evaluate generated images using alternative expert models Antelope v2 and Buffalo~\cite{insightface}, which are trained independently from the guidance model. While the Arcface model used for guidance exhibits higher similarities, all models agree that the average ID similarity is high enough (close to or higher than 0.5) for ID-preserved generation. This implies that our method does not dramatically overfit to a specific model’s behavior, but generates visually faithful images suitable for conditional generation.

\subsection{Working with advanced models}

\begin{wrapfigure}{r}{0.5\textwidth}
\vspace{-2em}
  \centering
  \includegraphics[width=0.5\textwidth]{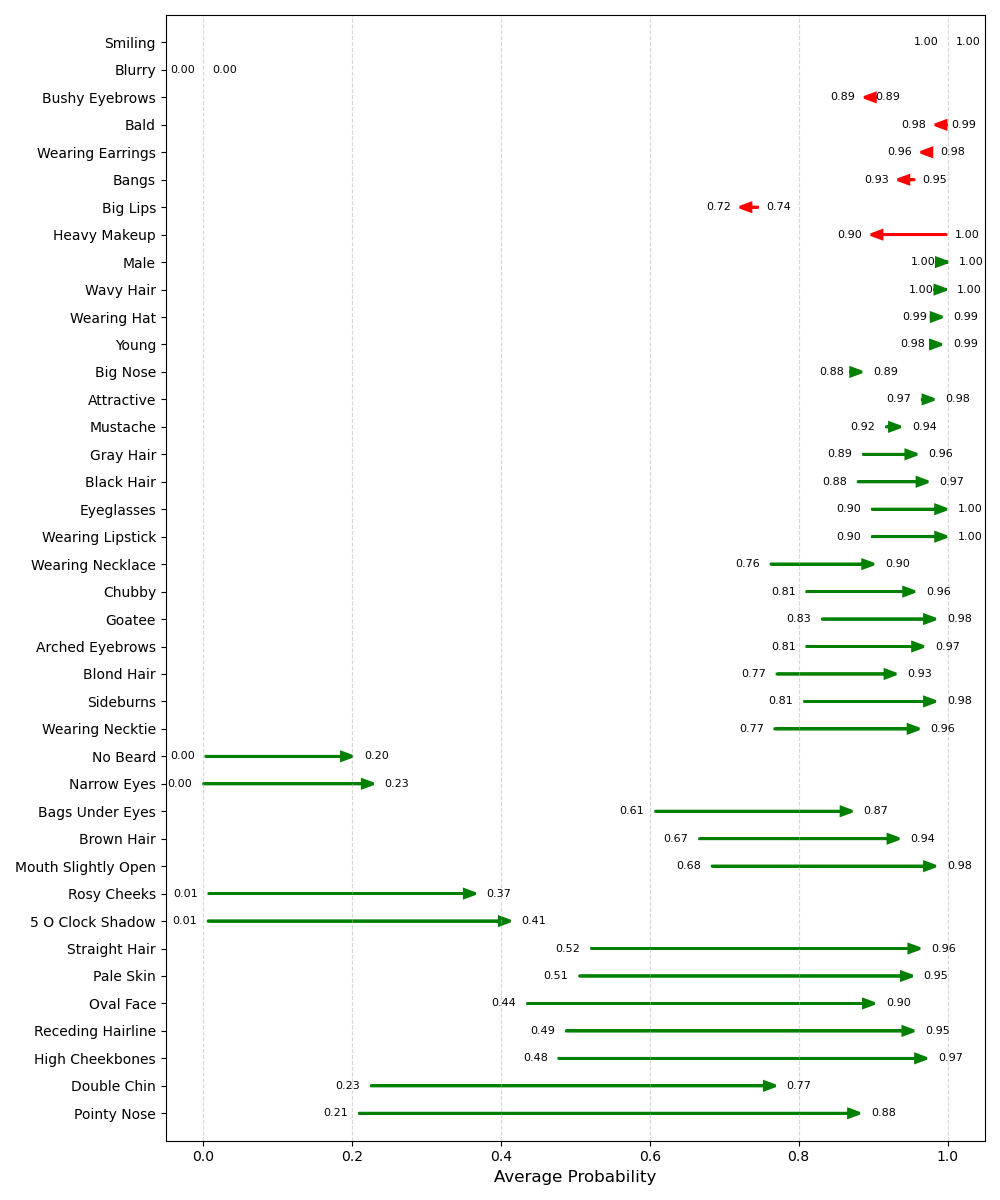}
  \caption{Average probability of successful generation on SDXL before and after ExpertGen across all 40 Celeb-A~\cite{celeba} facial attributes.}
  \label{fig:sdxl_attr}
\end{wrapfigure}

The proposed ExpertGen relies on enriched text prompts as an integral part of the method. This naturally leads to considerations about whether text conditions alone might be sufficient for certain tasks, particularly as more advanced models continue to emerge.

We show that larger models can still benefit from our training-free guidance methods. On the task of attribute guidance, SDXL~\cite{sdxl} achieves a promising performance of 72\% with text conditions alone, contrasting to only 48\% for SD-v1.5~\cite{ldm}. Nevertheless, ExpertGen achieves consistent improvement on SDXL, improving the average attribute accuracy of SDXL from 72\% to 85\% while maintaining high image quality. Per-attribute analysis in Fig.~\ref{fig:sdxl_attr} shows most attributes are improved to near-perfect levels. Gains are especially large for subtle attributes such as 5 o’clock shadows, receding hairlines, and pointy noses, which may suffer from limited paired training data. These results suggest that even state-of-the-art models may struggle with data sparsity in the long tail, which our expert guidance effectively mitigates.


\clearpage

\end{document}